\documentclass{article}

\usepackage[final,nonatbib]{neurips_2022}

\usepackage[utf8]{inputenc} 
\usepackage[T1]{fontenc}    
\usepackage{hyperref}       
\usepackage{url}            
\usepackage{booktabs}       
\usepackage{amsfonts}       
\usepackage{nicefrac}       
\usepackage{microtype}      
\usepackage{xcolor}         

\title{(Re)Defining Expertise in \\ Machine Learning Development}

\author{%
  Mark D\'iaz \\
  Google Research\\
  New York, NY 10011 \\
  \texttt{markdiaz@google.com} \\
  \And
  Angela D. R. Smith \\
  University of Texas at Austin \\
  Austin, TX 78712 \\
  \texttt{adrsmith@utexas.edu} \\
}

\begin{document}

\maketitle

\begin{abstract}
    Domain experts are often engaged in the development of machine learning systems in a variety of ways, such as in data collection and evaluation of system performance. At the same time, who counts as an ‘expert’ and what constitutes ‘expertise’ is not always explicitly defined. In this project, we conduct a systematic literature review of machine learning research to understand 1) the bases on which expertise is defined and recognized and 2) the roles experts play in ML development. Our goal is to produce a high-level taxonomy to highlight limits and opportunities in how experts are identified and engaged in ML research.
\end{abstract}

\section{Domain Experts in ML Development}
Human experts are engaged in various aspects of machine learning development, such as data collection and validation, consultation on the development of knowledge representations, and system evaluation. Moreover, who is identified as ‘expert’, and thus who partakes in these activities, carries implicit, normative claims about whose knowledge is valid and should be recognized as such. In ML, the label “expert” directs attention to whose knowledge should be upheld as canonical and used to shape how systems behave and how they are validated. In this way, expertise confers power and influence over system development to those identified as possessing it \cite{handy2007understanding}.

At the same time, definitions of domain expertise in ML vary, raising questions about who tends to to be recognized as an expert and why. Beyond the general notion of 'expertise' referring to a specialized or standard of knowledge, the bases on which expertise is identified can diverge. For example, in data annotation, “expert” has been used to refer to domain knowledge rooted in annotator lived experience \cite{patton2019annotating}, domain knowledge based on specific training processes (e.g., Wikipedia article editing) or academic study \cite{sen2015turkers}, and even to refer to labels from gold standard datasets, even when the knowledge or experience of gold standard annotators is unspecified \cite{snow2008cheap}. 

Scholars at the intersection of AI and HCI have expanded focus on the social dimensions of expertise \cite{park2021facilitating, abranches2019nurse}. For example, \cite{sambasivan2022deskilling} found that developers failed to recognize domain expertise, treated workers as non-essential, and de-skilled their work by treating them as data collectors. Complementing their investigations, we pursue a systematic review of ML publications to understand the nature and role of expertise in ML development. On the backdrop of recent critiques pointing out that expanded participation in ML does not necessarily improve equitable outcomes \cite{birhane2022power}, we consider not just \textit{what} kinds of expertise are recognized, but also \textit{how} experts participate.

\section{Systematic Literature Review}
Using \url{dblp.org}, we conducted a systematic literature review and thematic analysis \cite{braun2012thematic} of machine learning publications that involve domain experts in the development of a machine learning system. \url{dblp.org} is a bibliography of over 6 million computer science publications from a variety of venues, including prominent machine learning conferences such as NeurIPS, ICML, and AAAI, as well as broadly interdisciplinary venues, such as CHI and FAccT.

From an initial pre-search of ``expert',' ``expertise'', and ``domain expert'', we generated a list of search terms from the top-occurring bigrams, after removing stop-words. The list also included ``non-expert'', which we used to understand what did \textit{not} constitute expertise. In assessing relevant publications, we focused on papers that involved the development of a machine learning system, including papers that focused on components of ML systems, such as ontologies or causal maps. The publications excluded from our search results included entire books, PhD theses, magazine articles, review papers, and position papers in which no system was developed or evaluated. After removing papers that did not meet our inclusion criteria we were left with 96 papers.

For each paper we open coded 1) the bases for the definition of expertise used (e.g., expertise based on educational attainment) or if none was provided, we noted an implicit definition, 2) the phase(s) of development experts or non-experts contributed to (e.g., data collection), and 3) their role or how they participated (e.g., source data provider for ground truth annotations).

\section{Key Findings}
While the full code book is too extensive to present, we note two high-level findings:
\paragraph{1. Expertise is frequently unspecified} 38 of the 96 publications did not provide any explicit criteria that motivated the inclusion of the experts or non-experts engaged in system development or evaluation. In cases where the domain suggested a type of expert (e.g., medical experts for precision medicine systems) it was not clear why individual experts were selected or whether they possessed a particular expertise or focus within their field (e.g., cardiologist vs. ophthalmologist; years of practice). Failing to document explicit reasoning for determining expertise introduces issues for scientific reproducibility by under-specifying methodological details, as well as masks the reality of multiple ground truths \cite{aroyo2013crowd} which may be rooted in different cultural practices or training.  This also risks perpetuating representational harms \cite{barocas2017problem} by failing to acknowledge the range of expertise that may be relevant and by presenting an under-specified perspective as canonical. Explicitly naming the individuals or groups sought to influence system development and the basis for doing so is important for understanding which perspectives are represented within ML systems.



\paragraph{2. Experts were frequently engaged as data workers} Experts often provided or validated data annotations or provided term or concept definitions for ontology development. In 25 publications, experts annotated data or were themselves recorded as data subjects (e.g., using sensors). In only 2 publications was an expert involved in decisions about what a ML system should do or how an algorithm should perform (e.g., active decisions about what is included in an ontology, rather than providing domain definitions for pre-selected components). This points to the use of experts as a ``repository of knowledge'' \cite{handy2007understanding} that can regurgitate domain-specific facts, ignoring the breadth of skills that constitute expertise. Critiques of automation broadly have pointed out ways that it refocuses workers on menial tasks \cite{zuboff1988age}, rendering traditionally valued components of expertise, such as tacit knowledge, unimportant.

\section{Conclusion}
Ultimately, expertise should be explicitly documented and justified. In addition to supporting scientific reproducibility and transparency, there is a need to document relevant aspects of expertise to clarify whose knowledge is recognized and used to validate systems. From a responsible AI perspective, should also recognize and evaluate against expertise rooted in different cultural practices and lived experiences-- particularly as ML development increasingly turns to participatory approaches.

There is an additional need to engage expertise in responsible ways to minimize deskilling and incorporate knowledge beyond that which can be conveniently captured through data labeling. Much of ML development encompasses ‘participation as work’ \cite{sloane2020participation} leaving a range of expert participation minimally explored.

\bibliographystyle{abbrv}
\bibliography{neurips.bib}

\begin{thebibliography}{10}

\bibitem{abranches2019nurse}
D.~Abranches, D.~O'Sullivan, and J.~Bird.
\newblock Nurse-led design and development of an expert system for pressure
  ulcer management.
\newblock In {\em Extended Abstracts of the 2019 CHI Conference on Human
  Factors in Computing Systems}, pages 1--6, 2019.

\bibitem{aroyo2013crowd}
L.~Aroyo and C.~Welty.
\newblock Crowd truth: Harnessing disagreement in crowdsourcing a relation
  extraction gold standard.
\newblock {\em WebSci2013. ACM}, 2013.

\bibitem{barocas2017problem}
S.~Barocas, K.~Crawford, A.~Shapiro, and H.~Wallach.
\newblock The problem with bias: Allocative versus representational harms in
  machine learning.
\newblock In {\em 9th Annual conference of the special interest group for
  computing, information and society}, 2017.

\bibitem{birhane2022power}
A.~Birhane, W.~Isaac, V.~Prabhakaran, M.~D{\'\i}az, M.~C. Elish, I.~Gabriel,
  and S.~Mohamed.
\newblock Power to the people? opportunities and challenges for participatory
  ai.
\newblock {\em arXiv preprint arXiv:2209.07572}, 2022.

\bibitem{braun2012thematic}
V.~Braun and V.~Clarke.
\newblock {\em Thematic analysis.}
\newblock American Psychological Association, 2012.

\bibitem{handy2007understanding}
C.~Handy.
\newblock {\em Understanding organizations}.
\newblock Penguin Uk, 2007.

\bibitem{park2021facilitating}
S.~Park, A.~Y. Wang, B.~Kawas, Q.~V. Liao, D.~Piorkowski, and M.~Danilevsky.
\newblock Facilitating knowledge sharing from domain experts to data scientists
  for building nlp models.
\newblock In {\em 26th International Conference on Intelligent User
  Interfaces}, pages 585--596, 2021.

\bibitem{patton2019annotating}
D.~U. Patton, P.~Blandfort, W.~R. Frey, M.~B. Gaskell, and S.~Karaman.
\newblock Annotating twitter data from vulnerable populations: Evaluating
  disagreement between domain experts and graduate student annotators, 2019.

\bibitem{sambasivan2022deskilling}
N.~Sambasivan and R.~Veeraraghavan.
\newblock The deskilling of domain expertise in ai development.
\newblock In {\em CHI Conference on Human Factors in Computing Systems}, pages
  1--14, 2022.

\bibitem{sen2015turkers}
S.~Sen, M.~E. Giesel, R.~Gold, B.~Hillmann, M.~Lesicko, S.~Naden, J.~Russell,
  Z.~K. Wang, and B.~Hecht.
\newblock Turkers, scholars, "arafat" and "peace": Cultural communities and
  algorithmic gold standards.
\newblock In {\em Proceedings of the 18th ACM Conference on Computer Supported
  Cooperative Work \& Social Computing}, CSCW '15, page 826–838, New York,
  NY, USA, 2015. Association for Computing Machinery.

\bibitem{sloane2020participation}
M.~Sloane, E.~Moss, O.~Awomolo, and L.~Forlano.
\newblock Participation is not a design fix for machine learning.
\newblock {\em arXiv preprint arXiv:2007.02423}, 2020.

\bibitem{snow2008cheap}
R.~Snow, B.~O{'}Connor, D.~Jurafsky, and A.~Ng.
\newblock Cheap and fast {--} but is it good? evaluating non-expert annotations
  for natural language tasks.
\newblock In {\em Proceedings of the 2008 Conference on Empirical Methods in
  Natural Language Processing}, pages 254--263, Honolulu, Hawaii, Oct. 2008.
  Association for Computational Linguistics.

\bibitem{zuboff1988age}
S.~Zuboff.
\newblock {\em In the age of the smart machine: The future of work and power}.
\newblock Basic Books, Inc., 1988.

\end{thebibliography}

\end{document}